\documentclass{article} % For LaTeX2e
\usepackage{iclr2022_conference,times}

\usepackage{amsmath,amsfonts,bm}

\def\eqref#1{equation~\ref{#1}}
\def\1{\bm{1}}

\DeclareMathAlphabet{\mathsfit}{\encodingdefault}{\sfdefault}{m}{sl}
\SetMathAlphabet{\mathsfit}{bold}{\encodingdefault}{\sfdefault}{bx}{n}

\DeclareMathOperator*{\argmin}{arg\,min}

\usepackage{hyperref}
\usepackage{url}
\usepackage{booktabs}       % professional-quality tables
\usepackage{amsfonts}       % blackboard math symbols
\usepackage{nicefrac}       % compact symbols for 1/2, etc.
\usepackage{microtype}      % microtypography
\usepackage{xcolor}         % colors
\usepackage{amsmath}
\usepackage{multirow}
\usepackage{graphicx}
\usepackage{caption}
\usepackage{subcaption}
\usepackage{array}
\usepackage{wrapfig}
\usepackage{bbm}
\usepackage{inconsolata}
\usepackage{makecell}
\usepackage{comment}
\usepackage{fancyvrb,booktabs}
\usepackage{paralist}
\usepackage{tikz}
\usepackage{titlesec}

\newenvironment{tabularverbatim}
 {\VerbatimEnvironment
  \begin{BVerbatim}[baseline=c,formatcom=\setlength{\baselineskip}{\normalbaselineskip}]}
 {\end{BVerbatim}}

\titlespacing\section{0pt}{12pt plus 4pt minus 2pt}{0pt plus 2pt minus 2pt}
\titlespacing\subsection{0pt}{12pt plus 4pt minus 2pt}{0pt plus 2pt minus 2pt}
\titlespacing\subsubsection{0pt}{12pt plus 4pt minus 2pt}{0pt plus 2pt minus 2pt}

\title{Capturing Structural Locality in \\ Non-parametric Language Models}

\author{Frank F. Xu, Junxian He, Graham Neubig, Vincent J. Hellendoorn \\
School of Computer Science\\
Carnegie Mellon University\\
\texttt{\{fangzhex,junxianh,gneubig\}@cs.cmu.edu}, \texttt{vhellendoorn@cmu.edu}
}

\iclrfinalcopy % Uncomment for camera-ready version, but NOT for submission.
\begin{document}

\maketitle

\begin{abstract}
Structural locality is a ubiquitous feature of real-world datasets, wherein data points are organized into local hierarchies.
Some examples include topical clusters in text or project hierarchies in source code repositories.
In this paper, we explore utilizing this structural locality within \emph{non-parametric language models},
which generate sequences that reference retrieved examples from an external source.
We propose a simple yet effective approach for adding locality information into such models by adding learned parameters that improve the likelihood of retrieving examples from local neighborhoods.
Experiments on two different domains, Java source code and Wikipedia text, demonstrate that locality features improve model efficacy over models without access to these features, with interesting differences.
We also perform an analysis of how and where locality features contribute to improved performance and why the traditionally used contextual similarity metrics alone are not enough to grasp the locality structure.
\end{abstract}

\section{Introduction}

Language models (LMs) predict a probability distribution over sequences, and are most widely studied to model and generate natural languages~\citep{bengio2003neural,merity2018regularizing,baevski2018adaptive,brown2020language}.
% and to assess how algorithms understand human language~\citep{linzen2016assessing,kuncoro-etal-2017-recurrent}.
Advances in LMs benefit many natural language processing downstream tasks, such as machine translation~\citep{bahdanau2014neural}, dialog systems~\citep{sordoni2015neural}, question answering~\citep{yang2019xlnet,raffel2019exploring}, and general representation learning for natural language~\citep{devlin2018bert,liu2019roberta}.
Recently, LMs have also been adopted to model sequences other than text, such as source code written in programming language~\citep{hindle2016naturalness,hellendoorn2017deep,alon2020structural,karampatsis2020big}, which can enable useful downstream tasks like code completion~\citep{raychev2014code}.

Most current neural LMs are based on \emph{parametric} neural networks, using RNN~\citep{mikolov2010recurrent} or Transformer~\citep{vaswani2017attention} architectures.
These models make predictions solely using a fixed set of neural network parameters.
Recently, more and more neural LMs also incorporate \emph{non-parametric} components~\citep{grave2017unbounded,guu2018generating,he2020learning,Khandelwal2020Generalization}, which usually first select examples from an external source and then reference them during the prediction.
For example, \citet{Khandelwal2020Generalization} model the token-level probability by interpolating the parametric LM probability with a probability obtained from the nearest context-token pairs in an external datastore.
Using such non-parametric components in LMs is beneficial because the model no longer needs to memorize everything about the language in its parameters.

% \gn{The idea that ``distance'' is an important concept has not been introduced yet. Because this is important to understand the proposed method I think it's a good idea to introduce it somewhere, maybe in this paragraph. This clashes a little bit with ``some of these contexts may be inherently more useful than others'', as the distance is already a measure of this. I think maybe the argumentation should be something like the following (possibly reworded) ``Distance from the current context is a good measure of the likelihood that the context well be useful, but it is not perfect.''}

For such non-parametric LMs, one important concept is a \emph{distance} metric between the current context and other contexts in the datastore.
One example of such metric is the $\ell^2$ distance between context vectors calculated by the parametric model~\citep{Khandelwal2020Generalization}.
This distance can be used in both retrieval and probability calculation; items in the datastore that are less distant from the current context are more likely to be retrieved and have a higher influence on the final probability.
% They usually incorporate a distance metric as a function of two contexts, representing their similarity.
% 
However, given that non-parametric datastores are typically very large, containing a myriad of contexts from disparate sources, calculating a metric that accurately reflects semantic similarities is non-trivial; as we demonstrate in experiments, there is much room for improvement in current practice.
% We argue that distance from the current context is a good measure of the likelihood that the context will be useful, but it is not perfect. 
% Some of these contexts may be inherently more useful than others, in terms of helping predict the correct next token in LMs. 
% 
% \fx{Please check above.}
% \gn{The use of datastores in domain adaptation is also examined in \citet{Khandelwal2020Generalization}, right? Also in other papers on language modeling. I think it might be good to be clear what the big thing we're proposing here is.}.

\begin{figure}[!t]
\centering
    \vspace{-5mm}
    \includegraphics[width=0.8\textwidth]{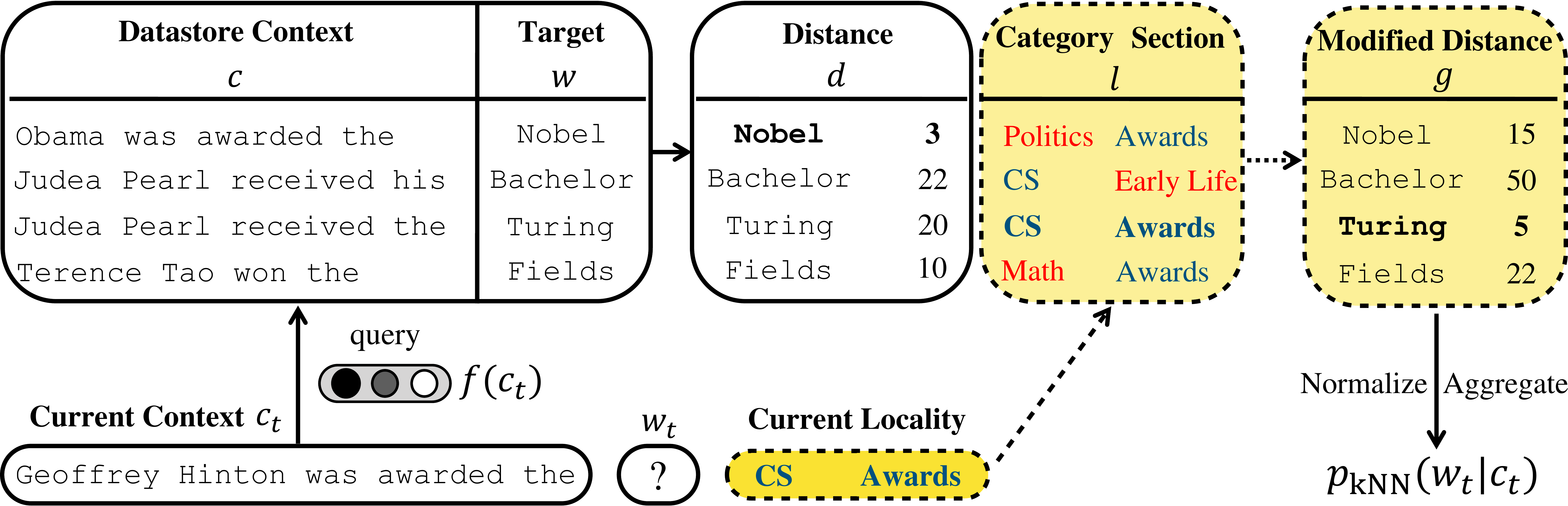}
    \caption{\label{fig:model}An example of incorporating structural locality in the computation flow of $p_{\text{kNN}}(w_t|c_t)$. The current context $c_t$ is used to calculate distance $d$ to contexts in the datastore $(c,w)$. Dashed boxes and lines represent components proposed in our work, which leverage structural information $l$ (\textcolor{red}{non-local}, \textcolor{blue}{local}) to allow for more accurate modified distances $g$ (lower is more similar).
    }
\vspace{-3mm}
\end{figure}

In this paper, we argue that the relevance of contexts may be correlated with not only contextual distance, but also structural characteristics of the underlying data.
Specifically, we take advantage of a property we dub \emph{structural locality}, the propensity of text to be divided into local groups sharing common hierarchical attributes.
This property is ubiquitous across many kinds of texts and can provide additional information on how closely related two different examples are to each other.
Throughout this paper, we will provide two case-studies of this phenomenon.
First, in the domain of programs written in source code, if two source files originate from the same project, they are more likely to be related than files from other projects, and even more so if they are from the exact same package~\citep{hellendoorn2017deep}.
Second, in natural language, two sections of Wikipedia text may be more related if they fall within the same topical domain, are from similarly titled sections, or even are from the same article (as in Figure~\ref{fig:model}).
Notably this locality often manifests itself at different levels, such as the levels of ``project'', ``subdirectory'', and ``file'' cited above for source code.

% An overview of how structural locality is used is shown in Figure~\ref{fig:model}. It highlights how a given context may match with a wide range of others on Wikipedia. Yet, we would expect that a section titled ``Awards'' from an article about computer scientists should be most closely related to, and thus helpful for, generating a section about another computer scientist's (here, Geoffrey Hinton) awards.

In this paper, we hypothesize that by using multiple levels of structural locality, we can better calibrate the distance metrics used to retrieve examples from non-parametric datastores, thereby improving LM performance.
% as the locality provides another aspect of ``relatedness'' of the contexts in the datastore, instead of only relying on the distances between the current context's and the retrieved contexts' representation.
Specifically, we propose a simple-yet-effective approach that can easily be applied to non-parametric LMs: we use different levels of structural locality to define functions that modify the contextual distance metrics used by the non-parametric module.

We evaluate our method on two drastically different domains: Java programming language source code, and natural language Wikipedia articles, achieving noticeable LM performance gains in both by adding just 5 \& 7 parameters respectively.
Moreover, we perform an in-depth analysis showing how the traditionally used contextual similarity metrics alone are not enough to grasp the locality structure, providing evidence for why adding the locality features is indeed useful.
We also compare programming languages and natural languages to highlight several interesting differences in terms of how, and how much, the locality helps improve LM performance.
 
\section{Non-parametric Language Models}
\label{sec:background}
Given a linguistic context consisting of a sequence of tokens $c_t = (w_1, . . . w_{t-1})$, autoregressive parametric LMs estimate $p(w_t|c_t;\theta)$, the probability distribution over the next token $w_t$.
Such parametric LMs store information regarding the language being modeled in the parameters $\theta$.
The size of $\theta$ is fixed in advance based on the hyperparameters of the model architecture, in recent years typically a neural network \citep{grave2016improving,baevski2018adaptive,dai2019transformer,brown2020language}.
In contrast, a non-parametric LM's number of parameters is not determined by just the model architecture, but also by the underlying data used to train the model.
While non-parametric LMs using Bayesian statistics have existed for some time~\citep{wood2011sequence,shareghi2017compressed,he2020learning},
they have recently seen increased prevalence through the introduction of neural LMs that retrieve relevant examples from an external datastore \citep{hashimoto2018retrieve,guu2018generating}.
In particular, we focus on kNN-LMs~\citep{Khandelwal2020Generalization}, a variety of such models that uses a nearest neighbor retrieval mechanism to augment a pre-trained parametric LM, achieving impressive results without any additional training.

Neural network-based LMs usually map the context $c$ to a fixed-length vector representation, with a trained function $f(c)$.
In kNN-LMs, the non-parametric component consists of a collection ($\mathcal{D}$) of contexts for the kNN to retrieve from.
Denoting these contexts and their corresponding next token as $(c_i, w_i) \in \mathcal{D}$, we create a datastore $(\mathcal{K}, \mathcal{V}) = \{(k_{i}, v_{i})\}$, which contains key-value pairs:
\begin{equation}
(\mathcal{K}, \mathcal{V})=\left\{\left(f\left(c_{i}\right), w_{i}\right) \mid\left(c_{i}, w_{i}\right) \in \mathcal{D}\right\}
\end{equation}

During inference, the parametric component of the LM generates the output distribution over next tokens $p_{LM}(w_t|c_t;\theta)$ and the corresponding context representation $f(c_t)$, given the test input context $c_t$. 
Then the non-parametric component of the LM queries the datastore with $f(c_t)$ representation
to retrieve its $k$-nearest neighbors $\mathcal{N}$ according to a distance function $d(\cdot,\cdot)$.
We can then compute a probability distribution over these neighbors using the softmax of their negative distances. 
The model aggregates the probability mass for each vocabulary item across all its occurrences in the retrieved targets. This distribution is then interpolated with
the parametric LM distribution $p_{\mathrm{LM}}$ to produce the final kNN-LM distribution:
\begin{align}
p_{\mathrm{kNN}}(w_t|c_t) &\propto \sum_{(k_{i}, v_{i}) \in \mathcal{N}} \textbf{1}_{w_t=v_{i}} \exp (-d(k_{i}, f(c_t)))\\\label{eq:pknn}
p(w_t|c_t;\theta) &= \lambda p_{\mathrm{kNN}}(w_t|c_t)+(1-\lambda) p_{\mathrm{LM}}(w_t|c_t;\theta)
\end{align}
In our experiments, we follow~\citet{Khandelwal2020Generalization} in setting the interpolation factor $\lambda$ to 0.25.

\section{Defining Structural Locality}
We define structural locality as a categorical feature calculated between a pair of contexts $(c_i, c_j)$ in a collection of data, that describes whether the pair share some common, potentially hierarchical, attributes (e.g., the section title of a Wikipedia article section, or the directory path of a source code file).
For each domain, a set of hierarchical attributes $\{l_0, l_1, ..., l_n\}$ can be defined based on prior knowledge of the domain.
We denote $l_k(c_i, c_j) \in \{0, 1\}$ as the boolean \emph{locality feature} value for the context pair, representing whether $c_i$ and $c_j$ share the same hierarchical attributes $l_k$.
Here, $l_0$ is reserved for ``no locality'', in case the pair shares none of the attributes.
Without loss of generality, we set a constraint that $\sum_k l_k(c_i, c_j) = 1 $, as new features can be introduced by conjunction and negation of the attributes if needed. % (as for Wikipedia text in Table \ref{tab:features}).

\vspace{0.5mm} \noindent \textbf{Specific Instantiations.}
We instantiate these features on our two case studies of Wikipedia text and Java source code, as summarized in Table~\ref{tab:features}.

\begin{table}[t]
\centering
\small
\vspace{-2mm}
\caption{Locality features designed for each data type according to domain knowledge.}
\label{tab:features}
\begin{tabular}{cll}
\toprule
Locality      & Wikipedia text                                 & Java projects                          \\ \midrule
$l_0$ & different article category, different section title & different project \\ \hline
$l_1$ & same article category, different section title & same project, different subdirectory \\ \hline
$l_2$ & same section title, different article category & same subdirectory                    \\ \hline
$l_3$ & same section title, same article category      & --                                    \\ \bottomrule
\end{tabular}
\vspace{-3mm}
\end{table}

% \gn{The following two paragraphs were in different sections but were repetitive. Merge them together.}
In Wikipedia, for every context $c_i$, we define four mutually exclusive hierarchical attributes, $l_0 - l_3$.
We calculate these features based on the Wikipedia article and section titles, using simple pattern matching.
We then link each article to a set of categories (one article may belong to multiple categories) using the knowledge graph WikiData,\footnote{\url{https://www.wikidata.org/}} by aggregating all the category entities involving two properties: P31 (instance of) and P279 (subclass of).
The criterion for ``same section title'' is exact string match~\citep{hayashi2020latent}.
If there is at least one common category between the sets of categories for two articles, the pair is assigned the ``same article category''.
% \gn{This description is not very precise, because in the case that both section title and article category are true then $l_1$, $l_2$, and $l_3$ would be true. It's evident what you mean if the reader thinks a bit ($l_1$ and $l_2$ are active only if $l_3$ is not), but it'd be better to state it explicitly}.

% \gn{The amount of description seems pretty imbalanced here compared to Wikipedia text. I think you can be a little bit more precise}
For Java source code, we define 3 mutually exclusive attributes, $l_0 - l_2$ based on the location of the code. For each source file, we use the full file path to obtain the two attributes: project name and sub-directory path.%
\footnote{For example, full path \texttt{.../Journal.IO/src/main/java/journal/io/api/DataFile.java} has project \texttt{Journal.IO} and sub-directory \texttt{src/main/java/journal/io/api/} for package \texttt{journal.io.api}.}
% In this case, the project name is \texttt{Journal.IO} and the sub-directory path is , corresponding to .
The criterion for both ``same project'' and ``same subdirectory'' is exact string match. Note that these features are strictly hierarchical, hence only two features are used to capture specific locality here.

\vspace{0.5mm} \noindent \textbf{An Aside: Connections to Domain Adaptation.}
% \gn{This is too much explanation, especially given that we're over the page limit. I'd take the first three paragraphs and compress them into one paragraph about the same length.}
Domain adaptation typically refers to reusing existing information about a given problem (e.g., data or model) to solve a task in a new domain.
Domain adaptation for neural models generally focuses on fine-tuning models on in-domain data~\citep{sennrich2016controlling,chu2017empirical} 
%by adding special domain tokens; 
or making direct modifications to the model to consider domain information~\citep{britz2017effective} or latent topic features~\citep{khudanpur2000maximum,mikolov2012context,wang2016larger}.
Most of these methods do not natively support new test-time contexts that were not seen at training time.
% For example, if you open a Java project in a code editor, and would like to adapt the Java source code LM to the current project, it would be costly to re-train or finetune on a project of files.
In comparison, one immediate advantage of non-parametric LMs is the ability to adapt to different domains at test time without re-training~\citep{merity2016pointer,grave2016improving,grave2017unbounded,Khandelwal2020Generalization}.
% Most of the methods make use of the context representation as they are crucial in neural LMs~\citep{khandelwal2018sharp}.
For example, some adaptive LMs~\citep{grave2016improving,grave2017unbounded} make use of the previous hidden states of test documents dynamically during inference.
Similarly, our proposed locality features do not require re-training on the training set.

% Other work attempts to make LMs more adaptive to different contexts by incorporating latent topic features~\citep{khudanpur2000maximum,mikolov2012context,wang2016larger}.
% Our proposed locality features differs in one key aspect: in prior work, the topic features are \emph{latent} distributions, while in our method, the locality features serve as an opportunity for the user to provide domain knowledge by specifying multiple levels of \emph{explicit} hierarchy attributes for a given dataset.
% This allows for more both interpretability and manual design of salient features.

Note that within the scope of this paper, although connected, the proposed \emph{structural locality} is a different concept from \emph{domain}.
We consider domains as higher-level classifications describing the text where one example belongs to one domain label; e.g., a section about Kim Kardashian's early life belongs to a category of texts describing celebrities.
One the other hand, with the structural locality, a user could define multiple levels of locality: to that same section, we can assign not only the domain label, but also, the section title ``Early Life''.
The lightweight nature of our model combined with non-parametric LMs also makes adding more levels of features straightforward, as the features only need to be calculated for the top nearest neighbors, and the number parameters that need tuning in our proposed method (Section \ref{sec:incorporating}) is only about twice the number of locality features.

% \fx{discuss related Cache Models like (Grave et al., 2017c; Merity et al., 2017) Sharp Nearby, Fuzzy Far Away?}

% \gn{Overall, as reading the intro, once we started talking about Wikipedia it made me think a bit more about how we frame this with respect to domain adaptation. I don't know of really good resources on domain adaptation in neural LMs, but here's a good survey from machine translation that might be a good place to start: \url{https://arxiv.org/pdf/1806.00258.pdf}. It might make the paper stronger if we could discuss the various approaches that exist for domain adaptation and what the difference/advantage of our approach is. I think a couple advantages are (1) like domain adaptation in the kNN-LM paper itself, our method doesn't require re-training, (2) it can capture multiple levels of locality. The paper might also be made stronger if you could add another baseline that captures locality in a different way, such as adding an embedding for each ``locality''.}
% \gn{Also, what is the difference between what we are calling ``structural locality'' and ``domain''? That's another thing that we should think about.}
% \fx{Please check above paragraph}

\section{Structural Locality and Nearest Neighbors}
\label{sec:knnanalysis}
In this section, we examine the relationship between distances derived from neural LM features $d(f(c_i),f(c_t))$, structural locality features $l(c_i, c_t)$, and the accuracy of the next-word prediction $w_i$.
Specifically, the underlying assumption of the kNN-LM is that less distant contexts will be more likely to accurately predict the next word $w_t$.
We would like to test whether this correlation between distance $d(\cdot)$ holds uniformly across different locality levels $l(\cdot)$, or if locality provides additional information indicative of whether a particular context is useful for predicting $w_i$ beyond just that provided by the neural representations.
% If the neural representations $f(c)$ are perfect and contexts local to the current context are indeed more relevant and thus closer in terms of their vector representation, then the distance metric should already include the information that our proposed locality features convey, rendering the features redundant.

\vspace{0.5mm} \noindent \textbf{Data.}
We use two different corpora from different domains to examine this question.

\textsc{Wikitext-103}\footnote{\url{https://blog.einstein.ai/the-wikitext-long-term-dependency-language-modeling-dataset/}.} is a standard language modeling benchmark~\citep{merity2016pointer} consisting of natural language text from English Wikipedia.
It contains a 250K token, word-level vocabulary, with 103M tokens in the training set and 250K tokens in both the validation and test sets.

\textsc{Java Github}\footnote{\url{https://zenodo.org/record/3628665}. } is a programming language corpus containing Java source code from Github~\citep{allamanis2013mining} that is widely used in source code modeling~\citep{hellendoorn2017deep,karampatsis2020big}.
It contains 1.44B tokens from 13,362 projects in the training split, 3.83M tokens from 36 projects in the validation split and 5.33M tokens from 38 projects in the test split.
The splits are separated by whole projects.
The dataset is tokenized with byte-pair encoding~\citep{sennrich2015neural} using the vocabulary from \citet{karampatsis2020big} with 2,000 subtokens.

\vspace{0.5mm} \noindent \textbf{Base Model.}
% Structural locality features are compatible for integration with many non-parametric retrieval based models as long as the model makes use of the distance between the retrieved candidates and the current context.
% The features will not affect the training of the underlying LM, as they are only used in inference.
As the neural model used to calculate context features, we follow \citet{Khandelwal2020Generalization},\footnote{\url{https://github.com/urvashik/knnlm}} train an LM with the exact architecture and optimization described by~\citet{baevski2018adaptive}: a decoder-only Transformer~\citep{vaswani2017attention}, with 1024 dimensional hidden states for the \textsc{Wikitext-103} dataset and 512 for \textsc{Java Github}.
We set the number of retrieved nearest neighbors to be analyzed to 1024, and the distance metric to $\ell^2$ following the default.

\vspace{0.5mm} \noindent \textbf{Datastore.}
% \gn{This explanation in this section sounds a bit convoluted. Couldn't we just say ``For WikiText, we include the training set, as well as the development/test set (excluding the file currently being evaluated) in the datastore. For the Java Github data, due to the relatively large size of the dev/set, and the unwieldy size of the training set, we include only the dev set (also excluding the current file).'' It could be a bit more verbose than this if there's something you really want to say, but this seems like enough?}
To capture the effect of our proposed locality features, the datastore should ideally be both closely related to the test examples, sufficiently large to ensure precise kNN retrieval performance for a wide range of contexts, and not too sparse in terms of the prevalence of locality features.

For \textsc{Wikitext-103}, we include the training set, as well as the validation/test set (excluding the text currently being evaluated) in the datastore. 
For the \textsc{Java Github}, due to the relatively large size of the validation/test set, and the unwieldy size of the training set, we include only the validation/test set (also excluding the current file).

% A datastore containing relevant contexts is required for the kNN retrieval.
% To study the effect of our proposed locality feature, it is ideal to have a datastore that is sufficiently large to not hurt the kNN retrieval performance, not too sparse in terms of the prevalence of locality features, and fair for the baselines' settings.

% For the \textsc{Java Github} dataset, we create the datastore by including the whole validation or test split depending on which split is being evaluated.
% Since the dataset is split by projects, this experiment choice allow us to have a datastore that contains files under the same directory or the same project, to verify the designed locality features.
% This dynamic setting is also closer to real world usage, simulating everyday development where programmers make small changes to existing code inside an opened project~\citep{karampatsis2020big}.\footnote{For comparing the effectiveness of locality we include other projects in the split as well.}
% The datastore is also sufficiently large for the kNN, as each split contains about 8,000 files spanning 40 projects.
% The kNN will only be able to retrieve contexts from the datastore \emph{excluding} the current test file itself to prevent cheating.

% For \textsc{Wikitext-103} dataset, in addition to the same choice as \textsc{Java Github}, we include the whole training set in the datastore to provide more contexts with denser locality features (e.g. more contexts under the same article category), as the validation or test set each only contains about 600 sections.

\begin{figure}[t]
\begin{subfigure}{1\textwidth}
  \centering
  \small
  \includegraphics[width=\linewidth]{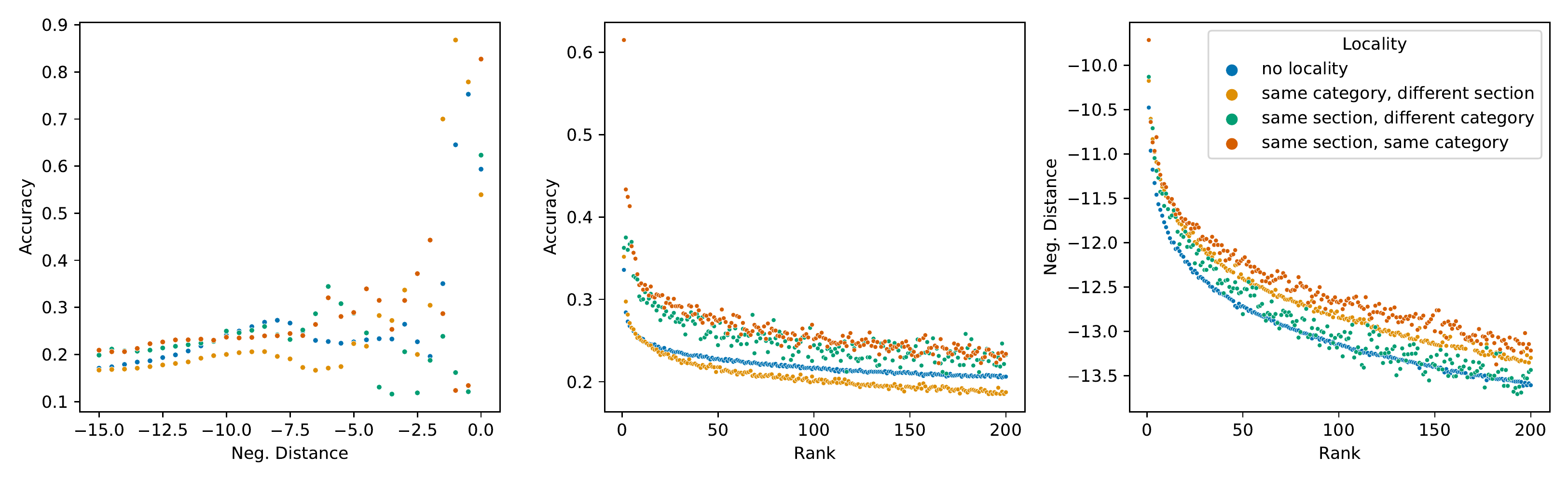}
  \caption{\textsc{Wikitext-103}}
  \label{fig:wikicorrelation}
\end{subfigure}\\%
\begin{subfigure}{1\textwidth}
  \centering
  \includegraphics[width=\linewidth]{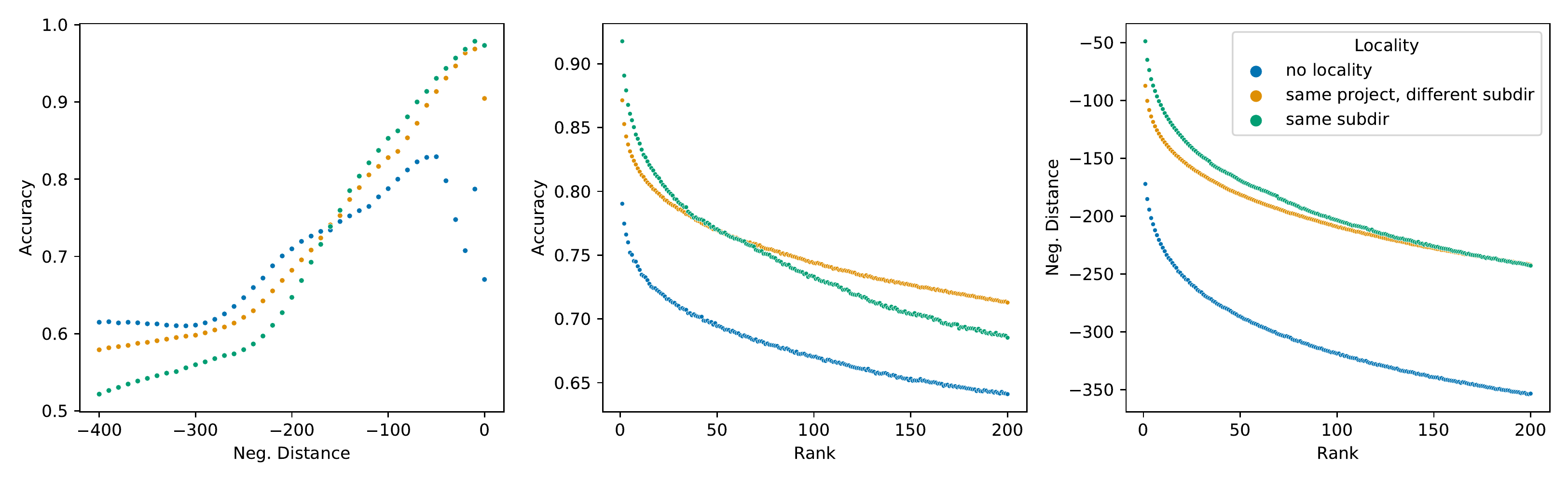}
  \caption{\textsc{Java Github}}
  \label{fig:javacorrelation}
\end{subfigure}
\caption{The relationship between nearest neighbor rank, negative distance, and the retrieval accuracy, grouped by different locality levels. Only top 200 nearest neighbors are shown for clarity. Negative distance on x-axis represents the upper bound of the bin.}
\vspace{-5mm}
\label{fig:correlation}
\end{figure}

\vspace{0.5mm} \noindent \textbf{Analysis.} 
% \gn{It would be good to more precisely define the quantities that you're measuring here, ideally with equations.}
% \fx{Please check below.}
Consider $k$ nearest neighbor contexts $\mathcal{N}_t = \{ c_r | r= 1...k\}$ retrieved for any test context $c_t$ in the test set $\mathcal{C}$, ordered by the ascending distance: $\forall r: d(c_r, c_t) < d(c_{r+1}, c_t)$.
We define $r \in [1, k]$ as the ``rank'' for the retrieved context $c_r$.
To study the quality of the retrieved contexts, we calculate the number of correctly retrieved tokens, defined as $\#\{w_r = w_{t_\text{gold}}\}$ across $\mathcal{C}$.

We plot in Figure~\ref{fig:correlation}, from left to right:
(1) Negative distances $\{-d(c_r, c_t) | c_r \in \mathcal{N}_t, c_t \in \mathcal{C}\}$ grouped into bins, vs. the retrieval accuracy of the bin $avg(\#\{w_r = w_{t_\text{gold}}\})$.
(2) Rank $r$ vs. the retrieval accuracy at rank $r$, $avg(\#\{w_r = w_{t_\text{gold}}\})$.
(3) Rank $r$ vs. the average negative distance $avg(-d(c_r, c_t))$ at rank $r$.
All of the plots are grouped by different locality levels $l_0$ to $l_n$.
% In Figure~\ref{fig:correlation}, we study the relationship between the nearest neighbor's rank among retrieved candidates, the negative distance, and the accuracy of the retrieved target token compared with the ground-truth token, grouped by different locality levels, on the test set.

Naturally, the left-most sub-figures reflect that the lower the (negative) distance, the lower the accuracy on both datasets.
Yet, interesting, on the Wikipedia domain (Figure~\ref{fig:wikicorrelation}), as the negative distance gets close to 0 (perfect match), the retrieval accuracy for the next word does not always increase; the accuracy values in this range have very high variance and all 4 levels of locality show no clear trend.
This partly indicates that context-based distance is imperfect, regardless of locality.
Even so, at slightly lower distances, the trends stabilize and largely show a consistent picture: more specific locality features, especially those involving the same category ($l_1 \& l_3$) yield better predictions than the locality-insensitive component for identical distances.
This is especially significant at higher ranked retrievals (middle sub-figure), where contexts that share the same section title and the same article category are substantially more likely to share the same completion.
This suggests that the proposed locality features are not fully represented by, or correlated with the original distance metric, and thus implies that there is room for improvement by incorporating these features.

In the Java source code domain (Figure~\ref{fig:javacorrelation}), we generally observe that the retrieval accuracy is much higher than that in the Wikipedia domain, suggesting that the kNN is doing a better job retrieving relevant contexts. 
This is largely due to higher repetitiveness of source code \citep{hindle2016naturalness}; as we show later, the base Transformer model also performs much better here than on natural language text (Section~\ref{sec:result}).
We also observe a more pronounced locality effect here: at the same distances close to 0 and at the same rank, neighbors that are local to the current context have far higher accuracy, indicating usefulness of locality features in the source code domain as well.
However, as we can see from the (right-most) plot of rank versus the negative distance, the average distances of the neighbors with higher locality levels are also significantly smaller than the distance of those without locality.
This suggests that the distance in the Java source code domain already correlates well with the level of locality, which may render incorporating locality features less beneficial. We study the precise benefit under one instantiation of this model next.
% This partly explains why the relative gain is smaller on \textsc{Java Github} after using locality features.

\section{Incorporating Structural Locality in Non-parametric LMs}
\label{sec:incorporating}

Now that we have demonstrated that locality is additionally indicative of next-word prediction accuracy beyond context distance, we propose a method to incorporate this information into the non-parametric retrieval module. % , guiding the likelihood for the pair of the current context and the retrieved context.
In the case of kNN-LMs (Section~\ref{sec:background}), recall that $p_{\mathrm{kNN}}$ is calculated based on the softmax of the negative distance $-d(f(c_i), f(c_t))$. % , where each $k_i = f(c_i)$ corresponds to a context $c_i$ in the datastore.
Assuming locality features $\{l_0, l_1, ..., l_n\}$ for each pair $(c_i, c_t)$ consisting the retrieved nearest neighbor and the current inference context $c_t$, we modify the formulation of $p_{\mathrm{kNN}}$ (Equation~\ref{eq:pknn}) to consider these features as below:
\begin{align}
p_{\mathrm{kNN}}(w_t|c_t;\{\theta_n\}) &\propto \sum_{(k_{i}, v_{i}) \in \mathcal{N}} \textbf{1}_{w_t=v_{i}} \exp (-g(k_i, c_t;\{\theta_n\}))\\
g(k_i, c_t;\{\theta_n\}) &= g_n(d(k_{i}, f(c_t)); \theta_n)~~       \mbox{if $l_n(c_i, c_t) = 1$}.\label{eq:learnable}
\end{align}
where $g_n(d(\cdot,\cdot); \theta_n)$ is a learnable function of the distance of the nearest neighbors, with parameter $\theta_n$ for each type of locality feature $l_n$.
One can view function $g(\cdot)$ as a ``modified'' distance for nearest neighbors after taking locality information into consideration.
In our experiments, we adopt a linear form of $g(\cdot)$:
\begin{equation}
    g_n(d(\cdot,\cdot); w_n, b_n) = w_nd(\cdot,\cdot) + b_n
\label{eqn:weight_bias}
\end{equation}
We omit the bias for $g_0(\cdot)$ by setting $b_0=0$ to remove one free parameter from the model and potentially make optimization easier.%
\footnote{We also experimented with an adaptive variant that conditioned the weights and biases ($\{w_n\},\{b_n\}$) on the current context representation $f(c_t)$ parameterized by a MLP. However, this did not result in significant improvement over directly optimizing $w$ and $b$ (Appendix~\ref{app:alternative_formulation}).}
To learn these functions, a user needs to provide just a small sample of annotated data in the same domain, as there are only $2n+1$ parameters to optimize.
In our experiments, we use the validation split for optimization.
The parameters are trained to minimize the negative log-likelihood of the kNN prediction of the gold token:
\begin{align}
    \argmin_{\{\theta_n\}} -\log p_{\mathrm{kNN}}(w_t=w_{t_\mathrm{gold}}|c_t; \{\theta_n\})
\end{align}
To optimize the parameters, we use the Adam~\citep{kingma2014adam} optimizer with a learning rate of 0.0001 on the validation set for 200 epochs. 
It converges within 20 minutes for both datasets.

\section{How Does Structural Locality Improve Language Modeling?}

\subsection{Experimental Setup}
\label{sec:setup}
\vspace{-1mm}

\vspace{0.5mm} \noindent \textbf{Baselines.}
Since we base our model on kNN-LMs, this model will be our most directly comparable baseline.
We also compare our model to the underlying parametric LM~\citep{baevski2018adaptive}, without the kNN module.
For the \textsc{Java Github} dataset, we additionally compare to the recent state-of-the-art model from \citet{karampatsis2020big} on code language modeling, which uses BPE and LSTMs.
In all experiments, the maximum number of tokens per input sample is 3,072.

\vspace{0.5mm} \noindent \textbf{Evaluation.}
We evaluate the performance of the LM with the standard perplexity metric and token prediction top-$k$ accuracy on the held-out data.%
\footnote{For \textsc{Java Github}, the perplexity is calculated on full tokens by aggregating the likelihood of subtokens. The accuracy is calculated that \emph{all} subtokens in a full token have to be predicted correctly.}
The top-$k$ accuracy is calculated by checking if the ground truth token is among the predicted top-$k$ list. This metric, primarily for $k \in \{1, 5\}$ (with more $k$ values in Appendix~\ref{app:additional_acc}), is commonly used to evaluate predictive models of source code \citep{hindle2016naturalness}.
In order to more easily incorporate and analyze the locality features, and also following~\citet{karampatsis2020big}, we split the evaluation dataset into independent test examples to evaluate, where each of the example is an atomic unit in the locality hierarchy.
For \textsc{Java Github}, each test example is a source code file, and for \textsc{Wikitext-103}, each example is an article section.\footnote{Note that because we predict \textsc{Wikitext-103} section-by-section instead of article-by-article the perplexity numbers reported here are somewhat worse than other works. Article-by-article calculation is not inherently incompatible with our proposed method, but it would require additional implementation to use different locality features for different locations in the output. Hence, we used section-by-section calculation for expediency.}

\begin{table}[t]
    \centering
    % \vspace{-0.3cm}
    \small
    \caption{Perplexity and top-$k$ token prediction accuracy results on two datasets. $^*$Uses released pre-trained model, $^\dagger$no stochastic training, for all others stddev $<$ 0.01 for 5 runs.}
    \label{tab:result}
    % \vspace{-0.1cm}
    % \small
    \begin{tabular}{llrrrrr}
    \toprule
   \textbf{Dataset} & \textbf{Model} & \textbf{Dev PPL} & \textbf{Test PPL} &  \makecell[r]{\textbf{Rel.} \\ \textbf{Gain}} & \makecell[r]{\textbf{Top-1 Acc.} \\ \textbf{(Rel. Err. Red.)}} & \makecell[r]{\textbf{Top-5 Acc.} \\ \textbf{(Rel. Err. Red.)}} \\
    \midrule
   \multirow{3}{*}{\makecell[l]{\textsc{Wikitext} \\ \textsc{~-103}}} & Transformer$^1$ & $^*$23.31 & $^*$23.73 & -- & 39.0\%\hfill(--) & 64.0\%\hfill(--) \\
    & +kNN$^2$ & $^\dagger$20.21 & $^\dagger$19.94 & 16.0\% & 41.3\%\hfill(3.79\%) & 66.8\%\hfill(7.91\%) \\
    & +kNN + locality & \textbf{19.51} & \textbf{19.16} & 3.9\% & \textbf{43.2\%}\hfill(3.29\%) & \textbf{68.0\%}\hfill(3.56\%) \\
    % & +kNN + dist. calibration  & 20.02 & 19.70 & -2.8\%  \\
    \midrule
    \multirow{4}{*}{\makecell[l]{\textsc{Java} \\ \textsc{Github}}} & BPE LSTM$^3$ & -- & $^*$5.27 & -- & -- & -- \\
    & Transformer & 3.29 & 3.07 & 41.7\% & 75.6\%\hfill(--) & 87.6\%\hfill(--) \\
    & +kNN & $^\dagger$2.43 & $^\dagger$2.18 & 29.0\% & 83.9\%\hfill(34.0\%) & 96.0\%\hfill(67.9\%) \\
    & +kNN + locality & \textbf{2.37} & \textbf{2.13} & 2.3\% & \textbf{84.7\%}\hfill(4.91\%) & \textbf{96.6\%}\hfill(15.0\%) \\
    % & +kNN + dist. calibration  & 2.41 & 2.17 & -1.9\%  \\
    \bottomrule
    \end{tabular}
    $^1$\cite{baevski2018adaptive}, $^2$\cite{Khandelwal2020Generalization}, $^3$\cite{karampatsis2020big}
    \vspace{-5mm}
\end{table}

\subsection{Results}
\label{sec:result}
\vspace{-1mm}
The high-level results are shown in Table~\ref{tab:result}.
At first glance, we can already see that the two datasets vary greatly in predictability.
With a similar Transformer architecture, performance on \textsc{Java Github} is much better than on the \textsc{Wikitext-103} across the board, partly due to the rigid nature of programming language syntax. With a Transformer model, we achieved a strong state-of-the-art language model on Java code, with low perplexity and very high prediction accuracy ({\raise.17ex\hbox{$\scriptstyle\sim$}}75\%).

By adding kNN module onto the Transformer-based LMs, perplexity and accuracy in both domains improves by a large margin.
This is expected and in line with previous experiments on kNN-LMs~\citep{Khandelwal2020Generalization}.
The Wikipedia domain enjoys less relative improvement in perplexity (16\%) than the Java source code domain (29\%).
This is particularly interesting, considering that the datastore used for \textsc{Wikitext-103} contains both the current held-out split and the training data ({\raise.17ex\hbox{$\scriptstyle\sim$}}100M contexts), compared to that for \textsc{Java Github} with only the current held-out split ({\raise.17ex\hbox{$\scriptstyle\sim$}}5M contexts).
This reflects the fact that source code is known to benefit strongly from project- and package-specific locality~\citep{tu2014localness,hellendoorn2017deep}.

\begin{wraptable}[10]{r}{0.4\textwidth}
\centering
\small
\vspace{-6mm}
\caption{Learned parameters $\theta_0, \{\theta_n\}$ for each locality level and a non-local level $g_0$.}
\label{tab:learnedparameters}
\vspace{-2mm}
\begin{tabular}{lcccc}
\toprule
& \multicolumn{2}{c}{\textsc{Wikitext-103}}  & \multicolumn{2}{c}{\textsc{Java Github}}\\
& $w$  & $b$ & $w$ & $b$\\ \midrule
$g_0$ & 1.233  & -- & 0.022 & -- \\ \hline
$g_1$ & 1.246 & -1.087 & 0.033 & -3.627 \\ \hline
$g_2$ & 1.288 & -1.250 & 0.041 & -5.920 \\ \hline
$g_3$ & 1.285 & -1.464 & -- & -- \\ \bottomrule
\end{tabular}
\end{wraptable}

Adding proposed locality features and finetuning the parameters on the validation set improves the performance further on both datasets, albeit with a smaller relative gain. This confirms our hypothesis that incorporating locality into the non-parametric retrieval-based LMs is beneficial.
We also see that locality features in the Wikipedia domain result in fairly consistent gains, while the Java source code domain sees especially strong accuracy improvements. 
This echoes our analysis of the source code corpus in Section~\ref{sec:knnanalysis}, where we found that distance was generally strongly correlated with accuracy, but that locality was particularly informative at low distances. There, it may help discriminate between top-ranked completion candidates (as also shown later in Tab. \ref{tab:case}).
It is notable that despite the fact that the perplexity and accuracy on \textsc{Java Github} are already very strong with the vanilla Transformer, we still see a noticeable relative error reduction of 4.9\% by adding locality levels information.

% Finally, we compare an ablation (``dist. calibration'') that only $g_0$ is optimized

\begin{figure}[t]
 \centering
\begin{subfigure}{0.4\textwidth}
  \centering
  \includegraphics[width=\linewidth]{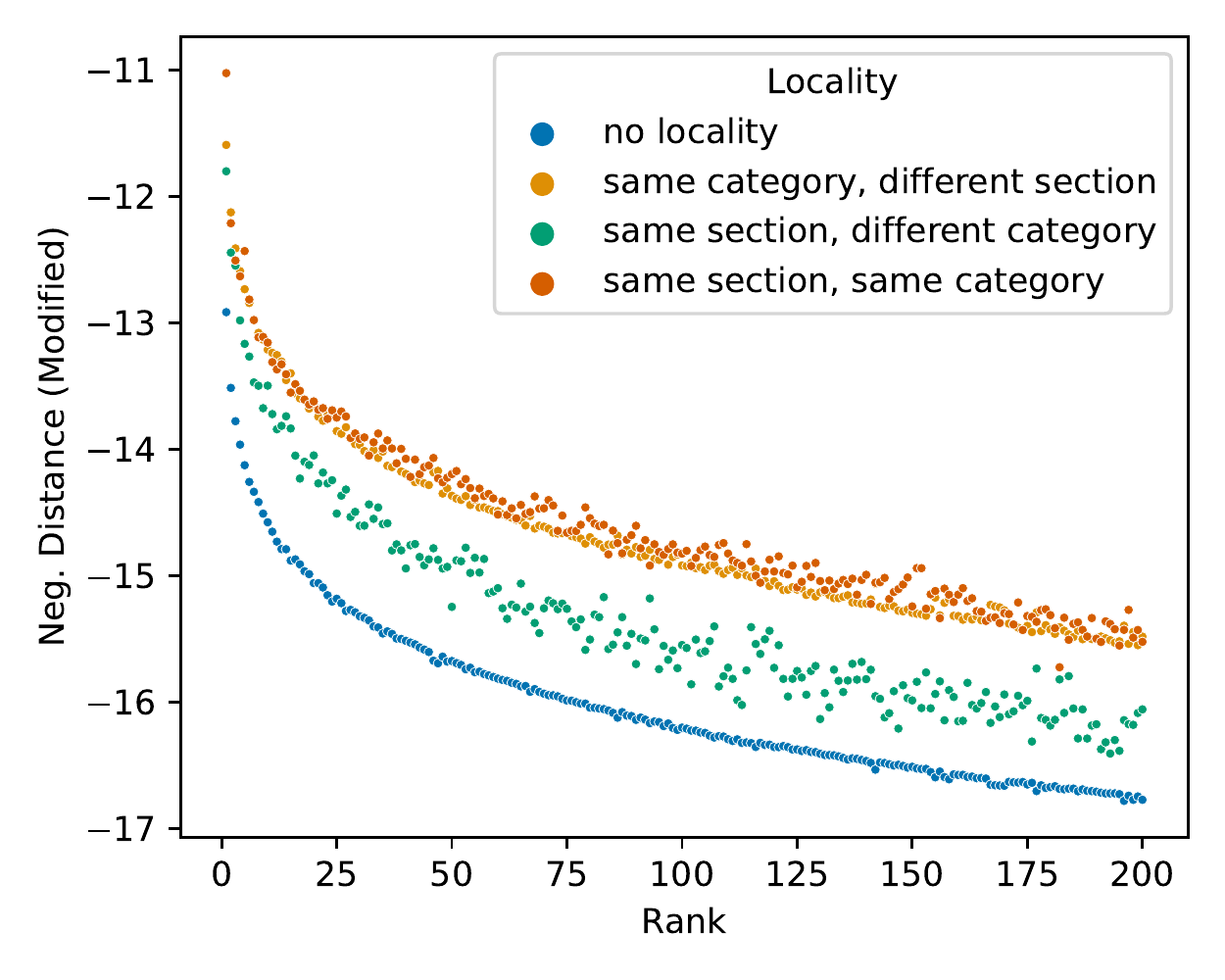}
  \caption{\textsc{Wikitext-103}}
  \label{fig:wikiafter}
\end{subfigure}\quad%
\begin{subfigure}{0.4\textwidth}
  \centering
  \includegraphics[width=\linewidth]{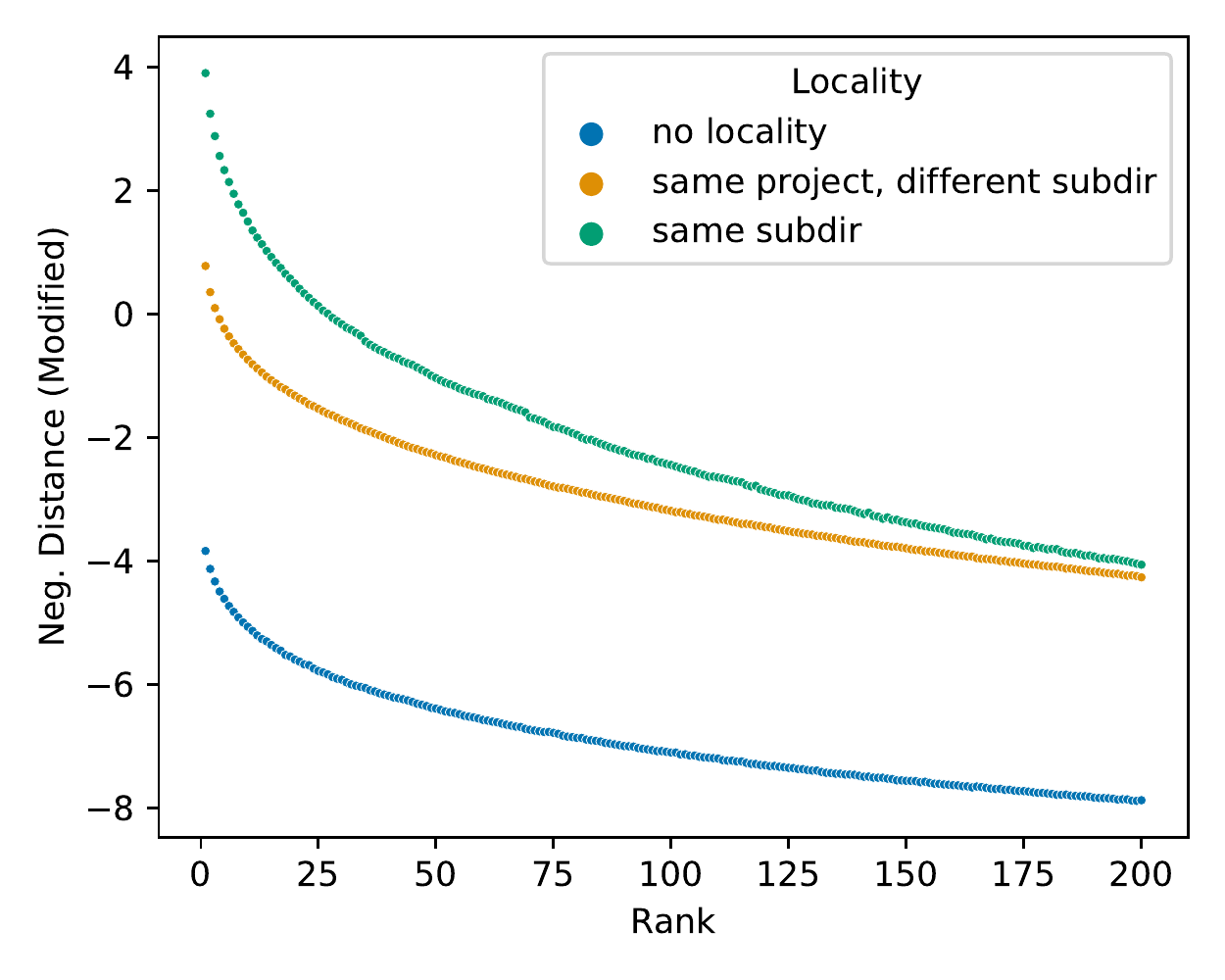}
  \caption{\textsc{Java Github}}
  \label{fig:javaafter}
\end{subfigure}
\caption{The relationship between nearest neighbor rank and the ``modified'' negative distance $-g$ guided by our proposed locality features, grouped by different locality levels. Similarly to Figure~\ref{fig:correlation}, only top 200 nearest neighbors are shown for clarity.}
\vspace{-5mm}
\label{fig:aftermodification}
\end{figure}

We next study how locality features guide towards a ``better'' distance distribution among nearest neighbors.
We plot the relationship between the nearest neighbor ranks and ``modified'' distance $g(k_i, c_t)$ in Figure~\ref{fig:aftermodification}.
Table~\ref{tab:learnedparameters} shows the specific learned parameters for each level of $g(\cdot, \cdot)$. Evidently, the biases and weights vary accordingly with locality levels, as the model tries to ``correct'' the original distance by emphasizing more local contexts.
Compared with the original negative distance $-d(k_i, f(c_t))$ depicted in Figure~\ref{fig:correlation}, the negative modified distance is more separated between the different locality levels on either dataset, showing the relative importance of different locality more clearly.
We analyze several alternative approaches to parameterizing locality in Appendix~\ref{app:alternative_formulation}.

For \textsc{Wikitext-103}, comparing Figure~\ref{fig:wikiafter} with Figure~\ref{fig:wikicorrelation}, we can see that with the original distance different localities cluster together, and the modified distance separates them much better.  
We can also see that if two contexts have the same article category and the same section title, then their distance on average is the closest, closely followed by those sharing article categories only.
On the other hand, contexts that only share section titles are not as closely related.
This is intuitively reasonable; the education section for a computer scientist and a musician can be very different in content, even if sharing some commonalities, like the phrases used to describe courses, grades, school locations, etc. This also underscores the usefulness of explicit locality features in providing interpretable insights into the relevance of domain knowledge.

For \textsc{Java Github}, comparing Figure~\ref{fig:javaafter} with Figure~\ref{fig:javacorrelation}, we can see that the original distance is already more separated between different locality levels than that of \textsc{Wikitext-103}, again suggesting better learned representations (in terms of locality sensitivity) for the Java domain.
However, the model still benefits somewhat from the contexts that are under the same subdirectory (more so than just the same project), especially for top nearest neighbors: the gap for ranks higher than 80 is more pronounced with the modified distance.
This again verifies our hypothesis about the hierarchical nature of structural locality.
It also indicates potential practical applications -- if this model were deployed in a code editor, one could obtain representations of the files in the same sub-directory as the current file and use them, along with the proposed locality features, to bias auto-complete results.

\begin{table}[t]
\centering
\caption{Examples from two domains where incorporating locality features (\textcolor{red}{non-local}, \textcolor{blue}{local}) lead to a significant increase in the cumulative $p_{\text{kNN}}$ for the gold token, with corresponding change in probability (normalized negative distance) for two nearest neighbors.}
\vspace{-3mm}
\small
\begin{tabular}{p{9cm}>{\centering\arraybackslash}m{1cm}>{\centering\arraybackslash}m{1.5cm}>{\centering\arraybackslash}m{1cm}}
        \toprule
\textbf{Test Context}         & \textbf{Test Target}        &    \textbf{Initial} $\log p_{\text{kNN}}$        &   $\Delta$ $\log p_{\text{kNN}}$      \\ \midrule
\textcolor{blue}{Section: Seasonal forecasts}; \textcolor{blue}{Category: Pacific typhoon season} \emph{The forecast indicated the potential for 26.2 tropical storms, compared to the 10– and 30-year average of 27.8 and 26.3 storms, respectively. The following month, the group raised their ...}& forecast &      -2.20        &  +0.89    \\\midrule
\textbf{Datastore Context} & \textbf{Datastore Target} & \textbf{Orig. Log-Prob.} &\textbf{$\Delta$Log-Prob.} \\ \midrule
\textcolor{blue}{Section: Seasonal forecasts}; \textcolor{blue}{Category: Pacific typhoon season} \emph{Their main reasons behind this is due to weaker trade force winds occurring in many parts of the basin, and there would be an enhanced cyclonic vorticity over the northwestern part of the Pacific. On April 27, the GCACIC made their first ...}  &  forecast     & -2.91  &  +1.25                  \\\addlinespace[0.5em]%\midrule
\textcolor{red}{Section: Earthquake}; \textcolor{red}{Category: earthquake} \emph{Nickson Sioni from Simbo came on the (HF) radio and announced the arrival of a huge wave that had washed away several houses and come inland about 200m. This information was passed on by telephone to the Hawaii-based Pacific Tsunami Warning Center who then upgraded their  ...}  &  warning    &  -3.01    &  -0.31                   \\
\bottomrule
\end{tabular}
\begin{tabular}{p{9cm}>{\centering\arraybackslash}m{1cm}>{\centering\arraybackslash}m{1.5cm}>{\centering\arraybackslash}m{1cm}}
        \toprule
\textbf{Test Context}         & \textbf{Test Target}        &    \textbf{Initial} $\log p_{\text{kNN}}$        &   $\Delta$ $\log p_{\text{kNN}}$      \\ \midrule
\textcolor{red}{Directory: .../android/twitter/AuthConstants.java}; \textcolor{blue}{Project: twitterdroid} 
\begin{tabularverbatim}
public static final String CONSUMER_SECRET = "YOUR_CONSUMER_SECRET";
public static final String REQUEST_URL = "http://www. ...
\end{tabularverbatim} 
& twitter &      -2.03        &  +0.49    \\\midrule
\textbf{Datastore Context} & \textbf{Datastore Target} & \textbf{Orig. Log-Prob.} &\textbf{$\Delta$Log-Prob.} \\ \midrule
\textcolor{red}{Directory: .../jtwitter/TwitterConnection.java}; \textcolor{blue}{Project: twitterdroid}
\begin{tabularverbatim}
public static final String FRIENDS_TIMELINE_URL =
"http://api.twitter.com/1/statuses/friends_timeline.xml";
public static final String UPDATE_URL = "http://api. ...
\end{tabularverbatim}
 &  twitter     & -1.99  &  +0.17  \\
\textcolor{red}{Directory: .../impl/ActivityTemplate.java}; \textcolor{red}{Project: spring-social-google} 
\begin{tabularverbatim}
private static final String ACTIVITIES_PUBLIC = "/activities/public";
private static final String ACTIVITIES_URL = "https://www. ...
\end{tabularverbatim}
  &  googleapis    & -1.87    &  -0.09                   \\
\bottomrule
\end{tabular}
\label{tab:case}
\vspace{-8mm}
\end{table}

Table~\ref{tab:case} shows a randomly sampled test context from each domain  where $p_{{\text{kNN}}}$ for the gold token increases after using locality features. We can see that the nearest neighbor search using context representations performs reasonably well at capturing patterns and themed phrases, especially closer to the last token, finding two very similarly rated candidates.
However, in both examples, the second retrieved candidate has a wrong target token. 
Informed by locality features -- in the \textsc{Wikitext-103} example, a matching section \emph{and} category for the first candidate -- the more ``local" context enjoys a large boost in probability, while the non-local one's decreases slightly.
We present additional examples in Appendix~\ref{app:additional_examples}.
The \textsc{Java} example demonstrates the same effect; the second retrieved example shows resemblances in variable name and template structures, but the fact that the project is focused on Google API rather than Twitter API makes the original retrieval undesirable.

\section{Conclusion}
\label{sec:conclusion}
In this paper, we propose a novel\footnote{Previous work~\citep{hellendoorn2017deep} made the observation that source code files from the same GitHub repository or sub-directory tend to be relatively similar, but did not include an empirical analysis of this effect. Rather, their observation was backed up by improved performance of their $n$-gram language model with multiple tiered caches. 
Our work gives more fine-grained insights into this phenomenon, expands the applicability to neural models and new domains, and proposes a more generalized formulation for encoding multiple localities across multiple domains. See Appendix~\ref{app:connection_related} for a detailed discussion of the connection to previous work and novelty.} 
method of incorporating structural locality into non-parametric LMs that reference retrieved examples from a datastore.
We evaluate this approach in both a natural language and programming language domain, and empirically explore the similarities and differences of how structural locality affects LMs in these settings.
The improvement in perplexity and prediction accuracy across both domains show the effectiveness and ubiquity of such locality information.
Besides language modeling, we also envision that the method can benefit other applications that could be enhanced using user-defined prior domain knowledge such as conditional generation or representation learning using retrieved information.

\vspace{0.5mm} \noindent \textbf{Limitations.}
Our method applies to settings where locality effects are present, there is sufficient data to build a reliable datastore for each locality level, and that locality is not already meaningfully captured by the model. While this may not apply to every domain, these features are common: besides source code \& Wikipedia, domains including books (features: authorship \& dewey decimal system information), research papers (venue, research area), product manuals (kind, sections), and online discussions (time, topic) are all plausible candidates.
% , given the performance improvement it returns for virtually no added parameters.
The features in our studied domains were selected based on availability and prior knowledge of the domain (e.g., for Java, \cite{hellendoorn2017deep}).
While they did provide measurable improvements and were natural to interpret, these may not be the optimal choice, and other options are worth investigating. 
It is also possible that LM improvements will eventually lead to learned context representations that almost perfectly capture the relevant locality information.
However, we believe this to be unlikely: in many practical settings, there is some inherent ambiguity in partial contexts that cannot be solved with the surface text only. For instance, in Java source code files, it is common to declare a package, which will obviously match perfectly based on the first few tokens (e.g., \texttt{package org.}) with many other contexts. Yet given the scoped nature of this declaration, locally retrieved continuations are inherently far more useful.

\section*{Acknowledgements}
We thank Uri Alon for the helpful discussions and thorough feedback.
We also thank the reviewers for the helpful conversations during the revision period of the paper. 
This work was supported in part by the National Science Foundation under Grant \#1815287, and a gift from Amazon AWS.

\section*{Ethics Statement}
There are several ethical considerations regarding our proposed method.
First, while language models have a large variety of positive uses in generating natural language \citep{li2021pretrained} or source code \citep{allamanis2018survey}, there is also a potential for dual use.
For example, previous work has cited the potential to generate fake news~\citep{zellers2019defending} or undesirable/defamatory content~\citep{wallace2019universal}.
Our methodology improves the accuracy of language models, which has the potential to increase their applicability not only in positive use cases but also in ethically questionable scenarios.
In order to mitigate these risks, methods to detect machine generated content may be employed, although these are not perfect remedies \citep{zellers2019defending}.

Second, because our methodology explicitly references the training corpus in generation it may increase the likelihood of copying content more-or-less verbatim from the training text.
This raises potential issues of copyright violation \citep{chen2021evaluating} or privacy violation \citep{carlini2020extracting}.
However, at the same time, because non-parametric models increase traceability through direct references to the training corpus, it also provides a tool to identify the provenance of the original content, providing a tool to identify such verbatim copying compared to standard parametric models.

Finally, there has been much recent interest in the energy and environmental impact of large language models \citep{strubell2019energy}.
Due to the necessity to look up information in a datastore, non-parametric models have additional computational overhead compared to parametric models.
However, at the same time, as noted by \citet{Khandelwal2020Generalization}, non-parametric models also provide a tool for quickly adapting to new domains through the use of domain-specific datastores, obviating the necessity for domain-specific fine-tuning.
Our work takes this a step further, allowing models to leverage locality of the datastores, potentially making this an even more attractive alternative for efficient adaptation.

\section*{Reproducibility Statement}
The source code package containing a README document on how to reproduce the results and analysis and experiment scripts is available in the paper's supplementary material. The details about the dataset used, model hyper-parameters, and analysis performed are described in Section~\ref{sec:knnanalysis} and Section~\ref{sec:setup}. All experiments are conducted on a single machine with a 48 core CPU and 8 NVIDIA V100 32GB GPU.
For \textsc{Wikitext-103} we use the pretrained model provided by~\citep{Khandelwal2020Generalization} for fair comparison.
For \textsc{Java Github} the Transformer model is trained until it converges, requiring approximately 2 days.
The datastore size is about 5GB for \textsc{Java Github} and 351GB for \textsc{Wikitext-103}.

\bibliography{ref}

\begin{thebibliography}{45}
\providecommand{\natexlab}[1]{#1}
\providecommand{\url}[1]{\texttt{#1}}
\expandafter\ifx\csname urlstyle\endcsname\relax
  \providecommand{\doi}[1]{doi: #1}\else
  \providecommand{\doi}{doi: \begingroup \urlstyle{rm}\Url}\fi

\bibitem[Allamanis \& Sutton(2013)Allamanis and Sutton]{allamanis2013mining}
Miltiadis Allamanis and Charles Sutton.
\newblock Mining source code repositories at massive scale using language
  modeling.
\newblock In \emph{2013 10th Working Conference on Mining Software Repositories
  (MSR)}, pp.\  207--216. IEEE, 2013.

\bibitem[Allamanis et~al.(2018)Allamanis, Barr, Devanbu, and
  Sutton]{allamanis2018survey}
Miltiadis Allamanis, Earl~T Barr, Premkumar Devanbu, and Charles Sutton.
\newblock A survey of machine learning for big code and naturalness.
\newblock \emph{ACM Computing Surveys (CSUR)}, 51\penalty0 (4):\penalty0 1--37,
  2018.

\bibitem[Alon et~al.(2020)Alon, Sadaka, Levy, and Yahav]{alon2020structural}
Uri Alon, Roy Sadaka, Omer Levy, and Eran Yahav.
\newblock Structural language models of code.
\newblock In \emph{International Conference on Machine Learning}, pp.\
  245--256. PMLR, 2020.

\bibitem[Baevski \& Auli(2018)Baevski and Auli]{baevski2018adaptive}
Alexei Baevski and Michael Auli.
\newblock Adaptive input representations for neural language modeling.
\newblock \emph{arXiv preprint arXiv:1809.10853}, 2018.

\bibitem[Bahdanau et~al.(2015)Bahdanau, Cho, and Bengio]{bahdanau2014neural}
Dzmitry Bahdanau, Kyunghyun Cho, and Yoshua Bengio.
\newblock Neural machine translation by jointly learning to align and
  translate.
\newblock In \emph{Proceedings of ICLR}, 2015.

\bibitem[Bengio et~al.(2003)Bengio, Ducharme, Vincent, and
  Jauvin]{bengio2003neural}
Yoshua Bengio, R{\'e}jean Ducharme, Pascal Vincent, and Christian Jauvin.
\newblock A neural probabilistic language model.
\newblock \emph{Journal of machine learning research}, 3\penalty0
  (Feb):\penalty0 1137--1155, 2003.

\bibitem[Britz et~al.(2017)Britz, Le, and Pryzant]{britz2017effective}
Denny Britz, Quoc Le, and Reid Pryzant.
\newblock Effective domain mixing for neural machine translation.
\newblock In \emph{Proceedings of the Second Conference on Machine
  Translation}, pp.\  118--126, 2017.

\bibitem[Brown et~al.(2020)Brown, Mann, Ryder, Subbiah, Kaplan, Dhariwal,
  Neelakantan, Shyam, Sastry, Askell, et~al.]{brown2020language}
Tom~B Brown, Benjamin Mann, Nick Ryder, Melanie Subbiah, Jared Kaplan, Prafulla
  Dhariwal, Arvind Neelakantan, Pranav Shyam, Girish Sastry, Amanda Askell,
  et~al.
\newblock Language models are few-shot learners.
\newblock \emph{arXiv preprint arXiv:2005.14165}, 2020.

\bibitem[Carlini et~al.(2021)Carlini, Tram\`er, Wallace, Jagielski,
  Herbert-Voss, Lee, Roberts, Brown, Song, Erlingsson, Oprea, and
  Raffel]{carlini2020extracting}
Nicholas Carlini, Florian Tram\`er, Eric Wallace, Matthew Jagielski, Ariel
  Herbert-Voss, Katherine Lee, Adam Roberts, Tom Brown, Dawn Song, \'Ulfar
  Erlingsson, Alina Oprea, and Colin Raffel.
\newblock Extracting training data from large language models.
\newblock In \emph{USENIX Security Symposium}, 2021.

\bibitem[Chen et~al.(2021)Chen, Tworek, Jun, Yuan, Ponde, Kaplan, Edwards,
  Burda, Joseph, Brockman, et~al.]{chen2021evaluating}
Mark Chen, Jerry Tworek, Heewoo Jun, Qiming Yuan, Henrique Ponde, Jared Kaplan,
  Harri Edwards, Yura Burda, Nicholas Joseph, Greg Brockman, et~al.
\newblock Evaluating large language models trained on code.
\newblock \emph{arXiv preprint arXiv:2107.03374}, 2021.

\bibitem[Chu et~al.(2017)Chu, Dabre, and Kurohashi]{chu2017empirical}
Chenhui Chu, Raj Dabre, and Sadao Kurohashi.
\newblock An empirical comparison of domain adaptation methods for neural
  machine translation.
\newblock In \emph{Proceedings of the 55th Annual Meeting of the Association
  for Computational Linguistics (Volume 2: Short Papers)}, pp.\  385--391,
  2017.

\bibitem[Dai et~al.(2019)Dai, Yang, Yang, Carbonell, Le, and
  Salakhutdinov]{dai2019transformer}
Zihang Dai, Zhilin Yang, Yiming Yang, Jaime~G Carbonell, Quoc Le, and Ruslan
  Salakhutdinov.
\newblock Transformer-{XL}: Attentive language models beyond a fixed-length
  context.
\newblock In \emph{Proceedings of ACL}, 2019.

\bibitem[Devlin et~al.(2018)Devlin, Chang, Lee, and Toutanova]{devlin2018bert}
Jacob Devlin, Ming-Wei Chang, Kenton Lee, and Kristina Toutanova.
\newblock B{ERT}: Pre-training of deep bidirectional transformers for language
  understanding.
\newblock \emph{arXiv preprint arXiv:1810.04805}, 2018.

\bibitem[Grave et~al.(2016)Grave, Joulin, and Usunier]{grave2016improving}
Edouard Grave, Armand Joulin, and Nicolas Usunier.
\newblock Improving neural language models with a continuous cache.
\newblock \emph{arXiv preprint arXiv:1612.04426}, 2016.

\bibitem[Grave et~al.(2017)Grave, Ciss{\'e}, and Joulin]{grave2017unbounded}
Edouard Grave, Moustapha Ciss{\'e}, and Armand Joulin.
\newblock Unbounded cache model for online language modeling with open
  vocabulary.
\newblock \emph{arXiv preprint arXiv:1711.02604}, 2017.

\bibitem[Guu et~al.(2018)Guu, Hashimoto, Oren, and Liang]{guu2018generating}
Kelvin Guu, Tatsunori~B Hashimoto, Yonatan Oren, and Percy Liang.
\newblock Generating sentences by editing prototypes.
\newblock \emph{Transactions of the Association for Computational Linguistics},
  6:\penalty0 437--450, 2018.

\bibitem[Hashimoto et~al.(2018)Hashimoto, Guu, Oren, and
  Liang]{hashimoto2018retrieve}
Tatsunori~B Hashimoto, Kelvin Guu, Yonatan Oren, and Percy~S Liang.
\newblock A retrieve-and-edit framework for predicting structured outputs.
\newblock In \emph{Proceedings of NeurIPS}, 2018.

\bibitem[Hayashi et~al.(2020)Hayashi, Hu, Xiong, and Neubig]{hayashi2020latent}
Hiroaki Hayashi, Zecong Hu, Chenyan Xiong, and Graham Neubig.
\newblock Latent relation language models.
\newblock In \emph{Proceedings of the AAAI Conference on Artificial
  Intelligence}, volume~34, pp.\  7911--7918, 2020.

\bibitem[He et~al.(2020)He, Berg-Kirkpatrick, and Neubig]{he2020learning}
Junxian He, Taylor Berg-Kirkpatrick, and Graham Neubig.
\newblock Learning sparse prototypes for text generation.
\newblock \emph{arXiv preprint arXiv:2006.16336}, 2020.

\bibitem[Hellendoorn \& Devanbu(2017)Hellendoorn and
  Devanbu]{hellendoorn2017deep}
Vincent~J Hellendoorn and Premkumar Devanbu.
\newblock Are deep neural networks the best choice for modeling source code?
\newblock In \emph{Proceedings of the 2017 11th Joint Meeting on Foundations of
  Software Engineering}, pp.\  763--773, 2017.

\bibitem[Hindle et~al.(2016)Hindle, Barr, Gabel, Su, and
  Devanbu]{hindle2016naturalness}
Abram Hindle, Earl~T Barr, Mark Gabel, Zhendong Su, and Premkumar Devanbu.
\newblock On the naturalness of software.
\newblock \emph{Communications of the ACM}, 59\penalty0 (5):\penalty0 122--131,
  2016.

\bibitem[Karampatsis et~al.(2020)Karampatsis, Babii, Robbes, Sutton, and
  Janes]{karampatsis2020big}
Rafael-Michael Karampatsis, Hlib Babii, Romain Robbes, Charles Sutton, and
  Andrea Janes.
\newblock Big code!= big vocabulary: Open-vocabulary models for source code.
\newblock In \emph{2020 IEEE/ACM 42nd International Conference on Software
  Engineering (ICSE)}, pp.\  1073--1085. IEEE, 2020.

\bibitem[Khandelwal et~al.(2020)Khandelwal, Levy, Jurafsky, Zettlemoyer, and
  Lewis]{Khandelwal2020Generalization}
Urvashi Khandelwal, Omer Levy, Dan Jurafsky, Luke Zettlemoyer, and Mike Lewis.
\newblock Generalization through memorization: Nearest neighbor language
  models.
\newblock In \emph{Proceedings of ICLR}, 2020.

\bibitem[Khudanpur \& Wu(2000)Khudanpur and Wu]{khudanpur2000maximum}
Sanjeev Khudanpur and Jun Wu.
\newblock Maximum entropy techniques for exploiting syntactic, semantic and
  collocational dependencies in language modeling.
\newblock \emph{Computer Speech \& Language}, 14\penalty0 (4):\penalty0
  355--372, 2000.

\bibitem[Kingma \& Ba(2014)Kingma and Ba]{kingma2014adam}
Diederik~P Kingma and Jimmy Ba.
\newblock Adam: A method for stochastic optimization.
\newblock \emph{arXiv preprint arXiv:1412.6980}, 2014.

\bibitem[Li et~al.(2021)Li, Tang, Zhao, and Wen]{li2021pretrained}
Junyi Li, Tianyi Tang, Wayne~Xin Zhao, and Ji-Rong Wen.
\newblock Pretrained language models for text generation: A survey.
\newblock \emph{arXiv preprint arXiv:2105.10311}, 2021.

\bibitem[Liu et~al.(2019)Liu, Ott, Goyal, Du, Joshi, Chen, Levy, Lewis,
  Zettlemoyer, and Stoyanov]{liu2019roberta}
Yinhan Liu, Myle Ott, Naman Goyal, Jingfei Du, Mandar Joshi, Danqi Chen, Omer
  Levy, Mike Lewis, Luke Zettlemoyer, and Veselin Stoyanov.
\newblock Roberta: A robustly optimized bert pretraining approach.
\newblock \emph{arXiv preprint arXiv:1907.11692}, 2019.

\bibitem[Merity et~al.(2016)Merity, Xiong, Bradbury, and
  Socher]{merity2016pointer}
Stephen Merity, Caiming Xiong, James Bradbury, and Richard Socher.
\newblock Pointer sentinel mixture models.
\newblock \emph{arXiv preprint arXiv:1609.07843}, 2016.

\bibitem[Merity et~al.(2018)Merity, Keskar, and Socher]{merity2018regularizing}
Stephen Merity, Nitish~Shirish Keskar, and Richard Socher.
\newblock Regularizing and optimizing {LSTM} language models.
\newblock In \emph{Proceedings of ICLR}, 2018.

\bibitem[Mikolov \& Zweig(2012)Mikolov and Zweig]{mikolov2012context}
Tomas Mikolov and Geoffrey Zweig.
\newblock Context dependent recurrent neural network language model.
\newblock In \emph{2012 IEEE Spoken Language Technology Workshop (SLT)}, pp.\
  234--239. IEEE, 2012.

\bibitem[Mikolov et~al.(2010)Mikolov, Karafi{\'a}t, Burget, {\v{C}}ernock{\`y},
  and Khudanpur]{mikolov2010recurrent}
Tom{\'a}{\v{s}} Mikolov, Martin Karafi{\'a}t, Luk{\'a}{\v{s}} Burget, Jan
  {\v{C}}ernock{\`y}, and Sanjeev Khudanpur.
\newblock Recurrent neural network based language model.
\newblock In \emph{Eleventh annual conference of the international speech
  communication association}, 2010.

\bibitem[Raffel et~al.(2019)Raffel, Shazeer, Roberts, Lee, Narang, Matena,
  Zhou, Li, and Liu]{raffel2019exploring}
Colin Raffel, Noam Shazeer, Adam Roberts, Katherine Lee, Sharan Narang, Michael
  Matena, Yanqi Zhou, Wei Li, and Peter~J Liu.
\newblock Exploring the limits of transfer learning with a unified text-to-text
  transformer.
\newblock \emph{arXiv preprint arXiv:1910.10683}, 2019.

\bibitem[Raychev et~al.(2014)Raychev, Vechev, and Yahav]{raychev2014code}
Veselin Raychev, Martin Vechev, and Eran Yahav.
\newblock Code completion with statistical language models.
\newblock In \emph{Proceedings of the 35th ACM SIGPLAN Conference on
  Programming Language Design and Implementation}, pp.\  419--428, 2014.

\bibitem[Sennrich et~al.(2015)Sennrich, Haddow, and Birch]{sennrich2015neural}
Rico Sennrich, Barry Haddow, and Alexandra Birch.
\newblock Neural machine translation of rare words with subword units.
\newblock \emph{arXiv preprint arXiv:1508.07909}, 2015.

\bibitem[Sennrich et~al.(2016)Sennrich, Haddow, and
  Birch]{sennrich2016controlling}
Rico Sennrich, Barry Haddow, and Alexandra Birch.
\newblock Controlling politeness in neural machine translation via side
  constraints.
\newblock In \emph{Proceedings of the 2016 Conference of the North American
  Chapter of the Association for Computational Linguistics: Human Language
  Technologies}, pp.\  35--40, 2016.

\bibitem[Shareghi et~al.(2017)Shareghi, Haffari, and
  Cohn]{shareghi2017compressed}
Ehsan Shareghi, Gholamreza Haffari, and Trevor Cohn.
\newblock Compressed nonparametric language modelling.
\newblock In \emph{IJCAI}, pp.\  2701--2707, 2017.

\bibitem[Sordoni et~al.(2015)Sordoni, Galley, Auli, Brockett, Ji, Mitchell,
  Nie, Gao, and Dolan]{sordoni2015neural}
Alessandro Sordoni, Michel Galley, Michael Auli, Chris Brockett, Yangfeng Ji,
  Margaret Mitchell, Jian-Yun Nie, Jianfeng Gao, and Bill Dolan.
\newblock A neural network approach to context-sensitive generation of
  conversational responses.
\newblock In \emph{Proceedings of NAACL}, 2015.

\bibitem[Strubell et~al.(2019)Strubell, Ganesh, and
  McCallum]{strubell2019energy}
Emma Strubell, Ananya Ganesh, and Andrew McCallum.
\newblock Energy and policy considerations for deep learning in nlp.
\newblock \emph{arXiv preprint arXiv:1906.02243}, 2019.

\bibitem[Tu et~al.(2014)Tu, Su, and Devanbu]{tu2014localness}
Zhaopeng Tu, Zhendong Su, and Premkumar Devanbu.
\newblock On the localness of software.
\newblock In \emph{Proceedings of the 22nd ACM SIGSOFT International Symposium
  on Foundations of Software Engineering}, pp.\  269--280, 2014.

\bibitem[Vaswani et~al.(2017)Vaswani, Shazeer, Parmar, Uszkoreit, Jones, Gomez,
  Kaiser, and Polosukhin]{vaswani2017attention}
Ashish Vaswani, Noam Shazeer, Niki Parmar, Jakob Uszkoreit, Llion Jones,
  Aidan~N Gomez, {\L}ukasz Kaiser, and Illia Polosukhin.
\newblock Attention is all you need.
\newblock In \emph{Proceedings of NeurIPS}, 2017.

\bibitem[Wallace et~al.(2019)Wallace, Feng, Kandpal, Gardner, and
  Singh]{wallace2019universal}
Eric Wallace, Shi Feng, Nikhil Kandpal, Matt Gardner, and Sameer Singh.
\newblock Universal adversarial triggers for attacking and analyzing nlp.
\newblock \emph{arXiv preprint arXiv:1908.07125}, 2019.

\bibitem[Wang \& Cho(2016)Wang and Cho]{wang2016larger}
Tian Wang and Kyunghyun Cho.
\newblock Larger-context language modelling with recurrent neural network.
\newblock In \emph{Proceedings of the 54th Annual Meeting of the Association
  for Computational Linguistics (Volume 1: Long Papers)}, pp.\  1319--1329,
  2016.

\bibitem[Wood et~al.(2011)Wood, Gasthaus, Archambeau, James, and
  Teh]{wood2011sequence}
Frank Wood, Jan Gasthaus, C{\'e}dric Archambeau, Lancelot James, and Yee~Whye
  Teh.
\newblock The sequence memoizer.
\newblock \emph{Communications of the ACM}, 54\penalty0 (2):\penalty0 91--98,
  2011.

\bibitem[Yang et~al.(2019)Yang, Dai, Yang, Carbonell, Salakhutdinov, and
  Le]{yang2019xlnet}
Zhilin Yang, Zihang Dai, Yiming Yang, Jaime Carbonell, Russ~R Salakhutdinov,
  and Quoc~V Le.
\newblock X{LN}et: Generalized autoregressive pretraining for language
  understanding.
\newblock In \emph{Proceedings of NeurIPS}, 2019.

\bibitem[Zellers et~al.(2019)Zellers, Holtzman, Rashkin, Bisk, Farhadi,
  Roesner, and Choi]{zellers2019defending}
Rowan Zellers, Ari Holtzman, Hannah Rashkin, Yonatan Bisk, Ali Farhadi,
  Franziska Roesner, and Yejin Choi.
\newblock Defending against neural fake news.
\newblock In \emph{Advances in Neural Information Processing Systems}, pp.\
  9051--9062, 2019.

\end{thebibliography}
\bibliographystyle{iclr2022_conference}

\newpage

\appendix
\section{Appendix}
\subsection{Additional Examples}
\label{app:additional_examples}
We provide additional examples on \textsc{Wikitext-103} in Table~\ref{fig:additional_case}.

\begin{table}[h]
\centering
\small
\begin{tabular}{p{9cm}>{\centering\arraybackslash}m{1cm}>{\centering\arraybackslash}m{1.5cm}>{\centering\arraybackslash}m{1cm}}
        \toprule
\textbf{Test Context}         & \textbf{Test Target}        &    \textbf{Initial} $\log p_{\text{kNN}}$        &   $\Delta$ $\log p_{\text{kNN}}$      \\ \midrule
\textcolor{blue}{Section: Design}; \textcolor{blue}{Category: ship class} \emph{In an effort to outmatch the American New York class, planners called for a ship armed with twelve 14-inch (36 cm) guns and faster than the 21 knots (39 km/h; 24 mph) of their rivals. Vickers files show that the Japanese had access to the designs for double- and triple-gun turrets, yet opted for six double turrets over four triple turrets. The final design—designated A-64 by the IJN—called for a ...}& displacement &      -2.54        &  +1.09    \\\midrule
\textbf{Datastore Context} & \textbf{Datastore Target} & \textbf{Orig. Log-Prob.} &\textbf{$\Delta$Log-Prob.} \\ \midrule
\textcolor{blue}{Section: Design}; \textcolor{blue}{Category: ship class} \emph{Both ships were also given torpedo bulges to improve their underwater protection and to compensate for the weight of the additional armour. In addition, their sterns were lengthened by 7.62 metres (25 ft). These changes increased their overall length to 213.8 metres (701 ft), their beam to 31.75 metres (104 ft 2 in) and their draft to 9.45 metres (31 ft). Their ...}  &  displacement     & -3.25  &  +1.23                  \\\addlinespace[0.5em]%\midrule
\textcolor{red}{Section: History}; \textcolor{red}{Category: gun mount} \emph{The British Admiralty ordered a prototype of Coles's patented design in 1859, which was installed in the ironclad floating battery, HMS Trusty, for trials in 1861, becoming the first warship to be fitted with a revolving gun turret. Coles's aim was to create a ...}  &  ship    &  -2.98    &  -0.24                   \\
\bottomrule
\end{tabular}
\begin{tabular}{p{9cm}>{\centering\arraybackslash}m{1cm}>{\centering\arraybackslash}m{1.5cm}>{\centering\arraybackslash}m{1cm}}
        \toprule
\textbf{Test Context}         & \textbf{Test Target}        &    \textbf{Initial} $\log p_{\text{kNN}}$        &   $\Delta$ $\log p_{\text{kNN}}$      \\ \midrule
\textcolor{red}{Section: La Venta}; \textcolor{blue}{Category: colossal statue} \emph{When discovered it was half-buried; its massive size meant that the discoverers were unable to excavate it completely. Matthew Stirling fully excavated the monument in 1940, after clearing the thick vegetation that had covered it in the intervening years. Monument 1 has been ...}& moved &      -2.97        &  +1.22    \\\midrule
\textbf{Datastore Context} & \textbf{Datastore Target} & \textbf{Orig. Log-Prob.} &\textbf{$\Delta$Log-Prob.} \\ \midrule
\textcolor{red}{Section: San Lorenzo}; \textcolor{blue}{Category: colossal statue} \emph{The sculpture suffered some mutilation in antiquity, with nine pits hollowed into the face and headdress. San Lorenzo Colossal Head 10 (also known as San Lorenzo Monument 89) has been ...}  &  moved     & -4.18  &  +1.36                  \\\addlinespace[0.5em]%\midrule
\textcolor{red}{Section: San Lorenzo}; \textcolor{red}{Category: castle} \emph{The excavations investigated the north of the fortress, searching for an entrance postulated by architect Eugene Viollet-le-Duc, but no such entrance was found. However, the excavation did reveal was that there was an addition to the north of the castle to enable the use of guns. Typologically, the structure has been ...}  &  dated    & -4.63    &  -0.11                   \\
\bottomrule
\end{tabular}

\caption{Additional \textsc{Wikitext-103} examples where incorporating locality features (\textcolor{red}{non-local}, \textcolor{blue}{local}) lead to a significant increase in the cumulative $p_{\text{kNN}}$ for the gold token, with corresponding change in probability (normalized negative distance) for two nearest neighbors.}
\vspace{-3mm}
\label{fig:additional_case}
\end{table}

\subsection{Additional Results on Token Prediction Accuracy}
\label{app:additional_acc}
We show additional results on top-$k$ ($k=10,20$) accuracy  and relative error reduction (RER) on two datasets in Table~\ref{tab:add_acc_result}.

\begin{table}[t]
    \centering
    % \vspace{-0.3cm}
    \small
    \caption{Additional token prediction top-$k$ ($k=10,20$) accuracy results and relative error reduction (RER) on two datasets.}
    \label{tab:add_acc_result}
    % \vspace{-0.1cm}
    \begin{tabular}{llrrrr}
    \toprule
   \textbf{Dataset} & \textbf{Model} & \textbf{Top-10} & \textbf{RER} & \textbf{Top-20} & \textbf{RER} \\
    \midrule
   \multirow{3}{*}{\textsc{Wikitext-103}} & Transformer & 72.0\% & - & 78.9\% & - \\
    & +kNN & 74.6\% & 9.29\% & 81.0\% & 9.98\% \\
    & +kNN + locality feat.  & \textbf{74.9\%} & 1.30\% & \textbf{81.1\%} & 0.84\%  \\
    \midrule
    \multirow{3}{*}{\textsc{Java Github}}& Transformer & 89.5\% & - & 90.8\% & - \\
    & +kNN & 97.3\% & 74.86\% & 98.2\% & 80.33\% \\
    & +kNN + locality feat.  & \textbf{97.9\%} & 21.89\% & \textbf{98.6\%} & 25.41\% \\
    \bottomrule
    \end{tabular}
 \end{table}

\subsection{Alternative Formulations to Learn Parameters for Locality Features}
\label{app:alternative_formulation}
An alternative way to incorporate locality features into the model is an adaptive variant that conditions the weights and biases ($\{w_n\},\{b_n\}$ in Equation~\ref{eqn:weight_bias}) on the current context representation $f(c_t)$ parameterized by a MLP:
\begin{equation}
    \begin{bmatrix}w_0 & ... & w_n &b_1 &... &b_n\end{bmatrix}^T = MLP(f(c_t))
\end{equation}

In our experiments, we used a two-layer MLP with ReLU activations, with 64 hidden units and 0.3 dropout rate during training. The perplexity results compared with directly optimizing weights and biases ($\{w_n\},\{b_n\}$) are shown in Table~\ref{tab:alt_mlp_result}.

We find that contextualizing the parameters does not result in significant improvements over directly optimizing $w$ and $b$, and sometimes makes the performance even worse. 
This is perhaps because the context vector space is very large (512-1024 dimensions) compared to the relatively few data points from the validation set used to train.

\begin{table}[t]
    \centering
    % \vspace{-0.3cm}
    \small
    \caption{The perplexity results comparing alternative formulation using MLP to contextualize parameters for locality features on two datasets.}
    \label{tab:alt_mlp_result}
    % \vspace{-0.1cm}
    \begin{tabular}{llrr}
    \toprule
   \textbf{Dataset} & \textbf{Model} & \textbf{Dev PPL} & \textbf{Test PPL}  \\
    \midrule
   \multirow{4}{*}{\textsc{Wikitext-103}} & Transformer & 23.31 & 23.73  \\
    & +kNN & 20.21 & 19.94  \\
    & +kNN + locality (MLP contextualized)  & 20.11 &  19.92 \\
    & +kNN + locality (direct)  & \textbf{19.51} &  \textbf{19.16} \\
    \midrule
    \multirow{4}{*}{\textsc{Java Github}}& Transformer & 3.29 & 3.07  \\
    & +kNN & 2.43 & 2.18  \\
    & +kNN + locality (MLP contextualized)  & 2.47 & 2.20 \\
    & +kNN + locality (direct)  & \textbf{2.37} & \textbf{2.13} \\
    \bottomrule
    \end{tabular}
 \end{table}

In Section~\ref{sec:result}, we discuss the effect of learned parameters for each locality level.
Observing that the bias terms ($b_i$) and weights ($w_i$)  vary according to the locality levels in the learned parameters and to study the weights of the non-local level $w_0$, we freeze all weights except for non-local weights ($w_{i>0}$) to 1 and only optimize bias terms and the weight for the non-local level ($w_0$).
This is to exacerbate the effect of bias on different locality levels.
The learned parameters are shown in Table~\ref{tab:additional_parameter_results}.
We see similar results where the bias terms vary aggressively to modify the ``distance'' with different levels of locality, and the weights for the non-local level are less than 1, lowering the importance of those non-local retrieved candidates.
It’s worth mentioning that for \textsc{Java Github} these learned biases are much larger in amplitude than before, to compensate for the small scale weights learned before (only around 0.03).
However, the perplexity results on both datasets are slightly worse than the full optimization setting that we use in the main experiments (19.33 vs. 19.16 in \textsc{Wikitext-103} and 2.15 vs. 2.13 in \textsc{Java Github}). 

\begin{table}[t]
    \centering
    % \vspace{-0.3cm}
    \small
    \caption{Learned parameters $\theta_0, \{\theta_n\}$ for each locality level and a non-local level $g_0$, with fixed $w_{i>0} = 1$ during optimization.}
    \label{tab:additional_parameter_results}
\begin{tabular}{lcccc}
\toprule
& \multicolumn{2}{c}{\textsc{Wikitext-103}}  & \multicolumn{2}{c}{\textsc{Java Github}}\\
& $w$  & $b$ & $w$ & $b$\\ \midrule
$g_0$ & 1.127  & -- & 0.901 & -- \\ \hline
$g_1$ & 1.000 &  -0.385 & 1.000 & -28.716 \\ \hline
$g_2$ & 1.000 & -0.475 & 1.000 & -55.428 \\ \hline
$g_3$ & 1.000 & -0.726 & -- & -- \\ \bottomrule
\end{tabular}

 \end{table}

\subsection{Connection with Related Work and Novelty}
\label{app:connection_related}
Previous work~\citep{hellendoorn2017deep} has made the observation that source code files from the same GitHub repository or sub-directory tend to be relatively similar, but did not include an empirical analysis of this effect. Rather, their observation was backed up by an improved performance of their $n$-gram language model with multiple tiered caches. 
Our work improves on this in a number of ways, including
\begin{inparaenum}
\item directly examining the internal representations of a neural language model, 
\item demonstrating that the internal representations do not sufficiently capture structural locality features,
\item providing efficient ways to compensate for this disconnect, leading to improved language modeling performance, and
\item showing that this carries over to Wikipedia, which has not been previously examined in this way. 
\end{inparaenum}
As a result, our work both gives more fine-grained insights into this phenomenon and expands the applicability to neural models and new domains. In addition, our work proposes a more generalized formulation for encoding multiple localities across multiple domains than the one proposed in \citet{hellendoorn2017deep}, which treated locality as strictly nested (e.g. project $\to$ sub-directory $\to$ file). Our formulation in Eq. \ref{eq:learnable} can encode more general hierarchies, such as the \emph{lattice} we used in the Wikipedia case:

\begin{center}
\begin{tikzpicture}[scale=.7]
  \node (top) at (0,2) {same section \& category};
  \node (a) at (-3,0) {same section};
  \node (b) at (3,0) {same category};
  \node (bot) at (0,-2) {any section \& category};
  \draw (bot) -- (a) -- (top) -- (b) -- (bot);
\end{tikzpicture}
\end{center}

\end{document}